\documentclass[conference]{IEEEtran}
\IEEEoverridecommandlockouts
\usepackage{cite}
\usepackage{amsmath,amssymb,amsfonts}
\usepackage{algorithmic}
\usepackage{graphicx}
\usepackage{textcomp}
\usepackage{xcolor}
\def\BibTeX{{\rm B\kern-.05em{\sc i\kern-.025em b}\kern-.08em
    T\kern-.1667em\lower.7ex\hbox{E}\kern-.125emX}}
\begin{document}

\title{SALAD: Self-Adaptive Lightweight Anomaly Detection for Real-time Recurrent Time Series\\
}

\author{\IEEEauthorblockN{Ming-Chang Lee}
\IEEEauthorblockA{\textit{Depart. of Information Security}\\
\textit{ and Communication Technology,}\\
\textit{Norwegian University of Science}\\ 
\textit{and Technology}\\
Gj{\o}vik, Norway \\
ming-chang.lee@ntnu.no}
\and
\IEEEauthorblockN{Jia-Chun Lin}
\IEEEauthorblockA{\textit{Depart. of Information Security}\\
\textit{ and Communication Technology,}\\
\textit{Norwegian University of Science}\\ 
\textit{and Technology}\\
Gj{\o}vik, Norway \\
jia-chun.lin@ntnu.no}
\and
\IEEEauthorblockN{Ernst Gunnar Gran}
\IEEEauthorblockA{\textit{Depart. of Information Security}\\
\textit{ and Communication Technology,}\\
\textit{Norwegian University of Science}\\ 
\textit{and Technology}\\
\textit{Simula Research Laboratory}\\
Gj{\o}vik, Norway \\
ernst.g.gran@ntnu.no}
}

\maketitle

\begin{abstract} \footnote{This is a draft preprint of a paper to be published in the 45th IEEE Computer Society Signature Conference on Computers, Software, and Applications (COMPSAC 2021), IEEE. The final paper may be different from this version. Please use the following citation for this paper:
Ming-Chang Lee, Jia-Chun Lin, and Ernst Gunnar Gran, “SALAD: Self-Adaptive Lightweight Anomaly Detection for Real-time Recurrent Time Series,” In Proceedings of the 45th IEEE Computer Society Signature Conference on Computers, Software, and Applications (COMPSAC 2021), IEEE.} 
Real-world time series data often present recurrent or repetitive patterns and it is often generated in real time, such as transportation passenger volume, network traffic, system resource consumption, energy usage, and human gait. Detecting anomalous events based on machine learning approaches in such time series data has been an active research topic in many different areas. However, most machine learning approaches require labeled datasets, offline training, and may suffer from high computation complexity, consequently hindering their applicability. Providing a lightweight self-adaptive approach that does not need offline training in advance and meanwhile is able to detect anomalies in real time could be highly beneficial. Such an approach could be immediately applied and deployed on any commodity machine to provide timely anomaly alerts. To facilitate such an approach, this paper introduces SALAD, which is a \underline{S}elf-\underline{A}daptive \underline{L}ightweight \underline{A}nomaly \underline{D}etection approach based on a special type of recurrent neural networks called Long Short-Term Memory (LSTM). Instead of using offline training, SALAD converts a target time series into a series of average absolute relative error (AARE) values on the fly and predicts an AARE value for every upcoming data point based on short-term historical AARE values. If the difference between a calculated AARE value and its corresponding forecast AARE value is higher than a self-adaptive detection threshold, the corresponding data point is considered anomalous. Otherwise, the data point is considered normal. Experiments based on two real-world open-source time series datasets demonstrate that SALAD outperforms five other state-of-the-art anomaly detection approaches in terms of detection accuracy. In addition, the results also show that SALAD is lightweight and can be deployed on a commodity machine.
\end{abstract}


\section{Introduction}
A time series is a series of data points that are evenly indexed in time order. Such time series are generated by various devices such as smart meters, system monitoring tools, IoT devices, etc. It is often the case that time series in the real world reflect a certain pattern that might occur periodically, such as transportation passenger volume, network traffic, click-through rate, energy usage, system resource consumption, and human gait. This kind of time series is referred as recurrent time series in this paper. Finding an anomalous event or abnormal behavior in time series has attracted great attention in recent years since it helps to trigger immediate reaction, prompt troubleshooting and enable appropriate countermeasures to be soon conducted.

During the last decade, a number of methods and approaches based on statistics or machine learning have been proposed for time series anomaly detection in different areas, including intrusion detection, system health monitoring, power demand monitoring, electrocardiogram monitoring, etc. However, statistical-based approaches require to obtain the target time series beforehand so as to generate a statistical model that fits the time series for conducting further anomaly detection. Due to this characteristic, statistical-based approaches might not be suitable to provide real-time anomaly detection on the fly (i.e., along with data collection). One might argue that statistical-based approaches can still be used on the fly. However, these approaches might generate inconsistent anomaly detection results as time goes by, meaning that they might produce different detection results over time (i.e., overwriting and denying some of anomalies that they have claimed previously). On the other hand, most approaches based on machine learning require humans to manually label training data, conduct an offline training process, or determine an appropriate detection threshold in advance. These requirements considerably limit the applicability and usefulness of these approaches in practice.

It is therefore highly valuable if an anomaly detection approach is able to learn the target time series without an offline training/learning process and able to accurately detect anomalies without the above-mentioned requirements. Furthermore, it might also be highly valuable to provide a lightweight anomaly detection approach that is able to be deployed on any commodity machine and able to detect anomalies in real time so that appropriate countermeasures can be undertaken as soon as possible.

To facilitate the abovementioned anomaly detection approach, this paper introduces SALAD, a  \underline{S}elf-\underline{A}daptive \underline{L}ightweight \underline{A}nomaly \underline{D}etection approach based on a special type of recurrent neural networks called Long Short-Term Memory (LSTM) \cite{hochreiter1997long}. SALAD is designed for recurrent time series data, especially for those that is generated in real time.  In order to keep computational cost as low as possible, SALAD employs LSTM with a simple network structure (e.g., one hidden layer with only 10 hidden nodes). This kind of LSTM is referred to as \emph{simple LSTM} in this paper. However, according to our experience, using a simple LSTM might not be able to effectively learn a recurrent time series or detect anomalies in the time series if the data pattern of the recurrent time series is complicated.

Therefore, instead of directly training a simple LSTM with the target time series to predict future data points and detect anomalies, SALAD converts the time series into a series of average absolute relative error (AARE) values on the fly. More specifically, SALAD keeps training a simple LSTM with historical data points in a sliding window manner to predict every upcoming data point and calculate the corresponding AARE value. Through the conversion, the resulting AARE series would be much smooth than the original time series, which helps both training and prediction.

When SALAD derives a sufficient number of AARE values, it starts detecting anomalies by training a simple LSTM with short-term historical AARE values to predict an AARE value for every upcoming data point. In the meanwhile, SALAD keeps updating its detection threshold based on all previously derived AARE values and the Three-Sigma Rule \cite{hochenbaum2017automatic}. If the difference between a predicted AARE value and its corresponding calculated  AARE value is higher than the self-adaptive detection threshold, the corresponding data point is considered anomalous. Otherwise, the data point is considered normal.

To evaluate SALAD, we used two real-world open-source time series datasets to compare SALAD with five state-of-the-art anomaly detection approaches. The results demonstrate that SALAD provides good detection accuracy in terms of precision, recall, and F-score. In addition, SALAD is time-efficient and cost-efficient, and it is able to directly run on a commodity machine. The contributions of this paper are as follows:
\begin{enumerate}
\item The proposed SALAD is a completely unsupervised anomaly detection approach without a requirement for offline training. It can be easily and immediately applied to a recurrent time series and detect anomalies on the fly without requiring human experts to pre-train a learning model, pre-build a data model, tune parameters, or to determine a detection threshold beforehand.
\item SALAD is a self-adaptive and flexible approach. It automatically re-trains its LSTM and adjusts its detection threshold to adapt to pattern changes in the target time series. These properties make SALAD an ideal anomaly detection approach for real-time recurrent time series. 
\item SALAD is lightweight and cost-effective. By converting a complex time series into a much simpler AARE series, SALAD successfully avoids using a deep LSTM structure, reduces computation complexity, and makes itself being able to deploy and execute on a commodity machine.
\end{enumerate}

The rest of the paper is organized as follows: Section II presents related work. In Section III, we introduce the details of SALAD. Section IV presents and discusses the experiments and the corresponding results. In Section V, we conclude this paper and outline future work.
\section{Related Work}

As stated earlier, anomaly detection approaches proposed in recent years could be statistical-based or machine learning-based. Statistical-based approaches work by establishing a statistical model for a given normal dataset and then use the model to check whether a new data point fits this model or not. If the data point has a low probability to be generated from the model, it is considered anomalous. For example, Twitter introduced two anomaly detection algorithms, called AnomalyDetectionTs (ADT for short) and AnomalyDetectionVec (ADV for short) \cite{hochenbaum2017automatic}. ADT is designed to detect statistically significant anomalies in a given time series, whereas ADV is designed to detect statistically significant anomalies in a given vector of observations without timestamp information. Since both ADT and ADV are statistical based, they need to obtain the target time series in advance for establishing the corresponding data model. Therefore, they might not be suitable to detect anomalies in a time series that is generated in real time. Furthermore, both ADT and ADV require human experts to set appropriate values to their parameters in order to achieve good detection performance. For example, it is required to set parameter \emph{max\_anoms} (which indicates the maximum number of anomalies to detect as a percentage of the time series) for both ADT and ADV. In addition, for ADV, it is also required to set parameter \emph{period} (which defines the number of data points in a single period, and used during seasonal decomposition) . 

luminol \cite{luminol} is proposed by LinkedIn and implemented as an open-source Python library for identifying anomalies in real user monitoring (RUM) data for LinkedIn pages and applications. Given a time series dataset, luminol calculates an anomaly score for each data point in the time series based on statistical analysis. If a data point has a high score, this data point is likely to be anomalous as compared with other data points in the time series. However, luminol relies on human experts to set a threshold to further determine which data points are anomalies. 

GrammarViz 3.0 is an anomaly detection approach for recurrent time series developed by Senin et al. \cite{senin2018grammarviz}. Given a recurrent time series, GrammarViz 3.0 discretizes the time series into a symbolic form, infers a context free grammar, and exploits its hierarchical structure to discover anomalies. Since GrammarViz 3.0 is statistical-based, it needs to obtain the whole time series in advance for offline analysis and consequently might not be suitable for detecting anomalies in a time series on the fly, i.e., along with the data acquisition. Although the authors have mentioned that GrammarViz 3.0 can process every upcoming data point in time, this approach announces anomalies until all data points in the time series have been processed. Therefore, it might not be able to trigger immediate reaction or enable appropriate countermeasures to be soon conducted. Furthermore, GrammarViz 3.0 requires human experts to set several parameters (e.g., \emph{window size}, \emph{Piecewise Aggregate Approximation size}, and \emph{alphabet size}) for each dataset.

Matrix Profile is a data structure and associated algorithms for time series analysis proposed by Yeh et al. \cite{yeh2016matrix}. Matrix Profile is able to discover motifs and find the top-\emph{K} number of discords where a motif is a repeated pattern in a time series and a discord is an anomaly. In order to use Matrix Profile, users are required to determine a window size, i.e., a user-defined subsequence length as mentioned in \cite{MatrixProfiles}. However, users need to decide an appropriate window size since it will significantly impact the execution result of Matrix Profile\cite{MatrixProfiles}. Furthermore, users also need to determine the value of \emph{K} such that Matrix Profile can report the top-\emph{K} anomalies that it discovers in a given time series. It means that this approach will always report \emph{K} anomalies it finds regardless of the actual number of anomalies in the time series. However, it would be difficult for users to configure this parameter since users cannot foresee the number of anomalies appear in a time series. Different from Matrix Profile, our proposed (i.e., SALAD) does not require users to determine a value for \emph{K} since SALAD is not based on the top-\emph{K} discord approach. The detection threshold used by SALAD is automatically calculated and updated at each time point to adapt to pattern changes in a given time series.     

Machine learning approaches are another category of anomaly detection. Most approaches belonging to this category require labelled datasets or/and offline training processes. For examples, Yahoo proposed EGADS \cite{laptev2015generic} to detect anomalies on time series based on a collection of anomaly detection and forecasting models. However, EGADS requires to go on an offline process to model the target time series before it can predict a data value later used by its anomaly detection module and its altering module. Lavin and Ahmad \cite{lavin2015evaluating} introduced Hierarchical Temporal Memory (HTM) for time series anomaly detection. However, HTM requires 15\% of a training dataset to be non-anomalous for training its neural network. Different from EGADS and HTM, our approach SALAD does not have these requirements.

Siffer et al. \cite{siffer2017anomaly} proposed a time series anomaly detection approach based on Extreme Value Theory. This approach makes no assumption on the distribution of time series and requires no threshold manually set by humans. However, this approach needs a long offline time period to do necessary calibration before it can conduct anomaly detection. Greenhouse \cite{lee2018greenhouse} is a zero-positive anomaly detection algorithm for time series based on LSTM. Greenhouse requires all data points in its training datasets to be non-anomalous, making Greenhouse a kind of supervised learning approach. During the training phase, Greenhouse adopts a Look-Back and Predict-Forward strategy to detect anomalies. For a given time point \emph{t}, a window of most recently observed values of length \emph{B} is used as \emph{Look-Back} to predict a subsequent window of values of length \emph{F} as \emph{Predict-Forward}. However, if the training data is not representative, Greenhouse might not be able to capture and accommodate pattern changes in the target time series.

RePAD \cite{lee2020repad} is a real-time anomaly detection approach for time series. RePAD utilizes a single LSTM model trained with short-term historical data points to predict and detect if any upcoming data point is anomalous or not. According to the experiment results shown in \cite{lee2020repad}, RePAD is able to detect anomalies either proactively or on time, but RePAD suffers from some undesirable false positives. To further reduce the false positive problem, Lee et al. \cite{lee2020rere} introduce a lightweight real-time ready-to-go anomaly detection approach called ReRe. ReRe employs two lightweight LSTM models with two levels of detection sensitivity to jointly detect if any upcoming data point is anomalous based on short-term historical data points. However, both RePAD and ReRe do not have ability to learn the repetitive pattern in a recurrent time series. In other words, they might not be able to recall the data pattern that they have seen before and might repeatedly generate false anomalies when processing subsequent recurrent time series. This phenomenon will be presented in Section IV. Different from RePAD and ReRe, SALAD does not have this issue since SALAD is designed to learn recurrent time series. SALAD has capability to remember recurrent patterns and detect anomalous data points that are significantly different from the data patterns.

\section{Methodology of SALAD}
When SALAD is launched to detect anomalies in a recurrent time series, SALAD firstly converts the target time series into a series of average absolute relative error (AARE) values. Such a process is called \emph{conversion}. Note that AARE is a well-known measure for determining the prediction accuracy of a forecast approach \cite{lee2020distributed}. Fig. 1 shows the conversion algorithm. Let \emph{t} be the current time point, and \emph{t} starts from 0 (i.e., the time point at which SALAD is launched). Let \emph{b} be the number of data points used to train a simple LSTM model in the conversion process. For example, \emph{b} = 100.
\begin{figure}[ht]
  \centering
  \includegraphics[width=0.5\textwidth]{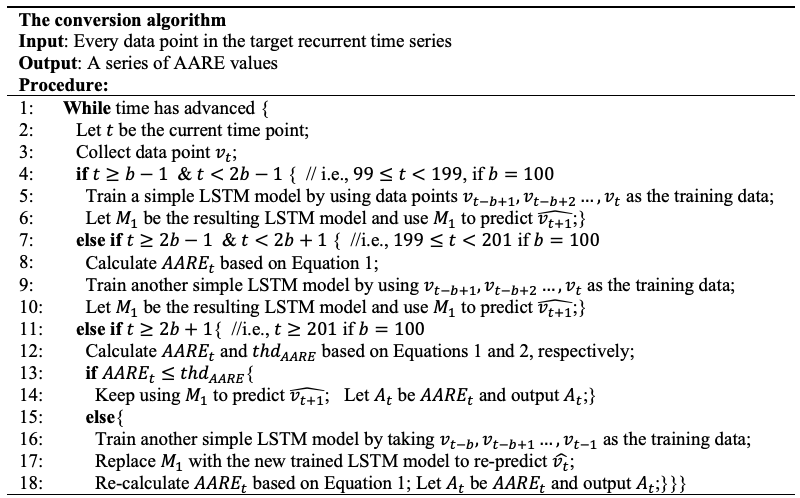}
  \caption{The conversion algorithm of SALAD.}
  \label{fig:Figure1}
\end{figure}

After SALAD is launched at time point 0, it keeps collecting every data point \emph{v\textsubscript{t}} observed at time point \emph{t} (see line 3 of Fig. 1). When \emph{t} has advanced to $\emph{b}$-1, SALAD uses all collected data points (i.e., \emph{v\textsubscript{t$-$b\emph{$+$1}},v\textsubscript{t$-$b\emph{$+$2}}, ..., v\textsubscript{t}}) to train a simple LSTM model (called \emph{M\textsubscript{\emph{1}}}) and then uses \emph{M\textsubscript{\emph{1}}} to predict the value of the next data point, denoted by $\widehat{\emph{v\textsubscript{t+1}}}$ (see lines 5 and 6 of Fig. 1). When \emph{t} further advances to \emph{2b-}1 (see line 7) and SALAD has collected \emph{v\textsubscript{b}}, SALAD is able to calculate an AARE value, denoted by \emph{AARE\textsubscript{t}}, based on the following equation:
\begin{equation}
AARE\textsubscript{\emph{t}}=\frac{1}{b} \cdot \sum_{y=t-b+1}^{t} \frac{\mid{v\textsubscript{\emph{y}}-\widehat{v\textsubscript{\emph{y}}}\mid}}{v\textsubscript{\emph{y}}}, \emph{t $\geq$ $2b-1$}
\end{equation}
where \emph{v\textsubscript{y}} is the observed data point at time point \emph{y}, and $\widehat{\emph{v\textsubscript{y}}}$ is the forecast data value at \emph{y}. For instance, if \emph{b} = 100 and \emph{t} = 199, $\emph{AARE\textsubscript{\emph{199}}}=\frac{1}{100}\cdot \sum_{y=100}^{199} \frac{\emph{$\mid${v\textsubscript{\emph{y}}}}-\widehat{\emph{v\textsubscript{\emph{y}}$\mid$}}}{\emph{v\textsubscript{\emph{y}}}}$. A lower AARE value indicates that the predicted value is close to the observed value. 

After calculating \emph{AARE\textsubscript{t}}, SALAD trains an simple LSTM model with the most recent \emph{b} data points (i.e., \emph{v\textsubscript{t$-$b\emph{$+$1}},v\textsubscript{t$-$b\emph{$+$2}}, ..., v\textsubscript{t}}), and then uses this model to predict the value of the next data point, denoted by  $\widehat{\emph{v\textsubscript{t\emph{$+$1}}}}$ (see lines 9 and 10). The abovementioned procedure will repeat while $\emph{2b}-1\leq\emph{t}<2\emph{b}+1$. 

As \emph{t} further advances to $2\emph{b}+1$ (see line 11), SALAD calculates \emph{AARE\textsubscript{t}} based on Equation 1 and then calculates a detection threshold called \emph{thd\textsubscript{AARE}} based on Equation 2 (see line 12) by considering all previously derived AARE values and following the Three-Sigma Rule \cite{hochenbaum2017automatic}, commonly used rule for anomaly detection.
\begin{equation}
thd\textsubscript{\emph{AARE}}=\mu\textsubscript{\emph{AARE}}+3\cdot\sigma
\end{equation}
In Equation 2, $\mu\textsubscript{\emph{AARE}}=\frac{1}{t-b-1}\cdot \sum_{x=2b-1}^{t} \emph{AARE\textsubscript{x}}$, and $\sigma$ is the standard deviation derived as below. 
\begin{equation}
\sigma=\sqrt{\frac{\sum_{x=2b-1}^{t}{(AARE\textsubscript{\emph{x}}-\mu\textsubscript{\emph{AARE}})^2}}{t-b-1}}
\end{equation} 
As shown from lines 13 to 14, if \emph{AARE\textsubscript{t}} is smaller than or equal to \emph{thd\textsubscript{AARE}}, implying that \emph{v\textsubscript{t}} is similar to previous data points, SALAD uses the same LSTM model (i.e., \emph{M\textsubscript{\emph{1}}}) to predict the next data point $\widehat{\emph{v\textsubscript{t+1}}}$. In addition, SALAD renames \emph{AARE\textsubscript{t}} as \emph{A\textsubscript{t}} and outputs \emph{A\textsubscript{t}} immediately.

However, if \emph{AARE\textsubscript{t}} is larger than \emph{thd\textsubscript{AARE}} (see line 15), implying that the data pattern of the target time series has changed, SALAD trains another  simple LSTM model to replace \emph{M\textsubscript{\emph{1}}} by taking the most recent \emph{b} data points (i.e., \emph{v\textsubscript{t$-$b},v\textsubscript{t$-$b\emph{$+$1}}, ..., v\textsubscript{t$-$\emph{1}}}), as the training data. The new trained LSTM model is immediately used to re-predict $\widehat{\emph{v\textsubscript{t}}}$. The purpose is to adapt to the pattern change and to make further prediction as accurate as possible. After that, the corresponding \emph{AARE\textsubscript{t}} is re-calculated, renamed as \emph{A\textsubscript{t}}, and then outputted (see lines 16 and 18).

Fig. 2 illustrates the detection algorithm of SALAD. When \emph{t} has advanced to time point 2\emph{b$+$\emph{1}}, SALAD starts the detection algorithm by inputting every single calibrated AARE value (denoted by \emph{A\textsubscript{t}}) from the conversion algorithm. Before SALAD can officially detect anomalies in the target time series, it needs to obtain three historical AARE values from the conversion algorithm so as to predict the next AARE value. As shown in Fig. 2, the detection algorithm requires to firstly collect \emph{A\textsubscript{$2b+$\emph{1}}}, \emph{A\textsubscript{$2b+$\emph{2}}}, and \emph{A\textsubscript{$2b+$\emph{3}}} at time points $\emph{$2b+$\emph{1}}$, $\emph{$2b+$\emph{2}}$, and $\emph{$2b+$\emph{3}}$, respectively. At time point $\emph{$2b+$\emph{3}}$, SALAD is able to train a simple LSTM model by taking \emph{A\textsubscript{$2b+$\emph{1}}}, \emph{A\textsubscript{$2b+$\emph{2}}}, and \emph{A\textsubscript{$2b+$\emph{3}}} as the training data. The trained LSTM model, denoted by \emph{M\textsubscript{\emph{2}}}, is then immediately used to predict the AARE value of the next time point, denoted by $\widehat{\emph{A\textsubscript{$2b+$\emph{4}}}}$ (see lines 3 to 5 of Fig. 2).
\begin{figure}[ht]
  \centering
  \includegraphics[width=0.5\textwidth]{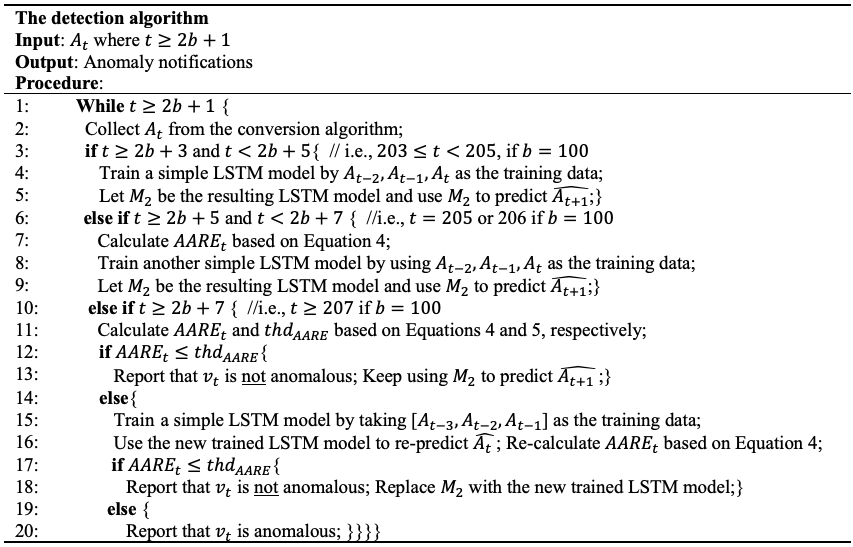}
  \caption{The detection algorithm of SALAD. }
  \label{fig:Figure2}
\end{figure}

When time has advanced to \emph{$2b+5$} (see line 6), SALAD calculates \emph{AARE\textsubscript{$2b+5$}} according to the following equation:

\begin{equation}
AARE\textsubscript{\emph{t}}=\frac{1}{3} \cdot \sum_{z=t-2}^{t} \frac{\mid{A\textsubscript{\emph{z}}-\widehat{A\textsubscript{\emph{z}}}\mid}}{A\textsubscript{\emph{z}}}, \emph{t $\geq$ $2b+5$}
\end{equation}

In Equation 4, \emph{z} is the index of time where \emph{z $\geq$ $2b+3$}, \emph{A\textsubscript{z}} is the AARE value calibrated and output by the conversion algorithm for time point \emph{z}, and $\widehat{\emph{A\textsubscript{z}}}$ is the AARE value predicted by \emph{M\textsubscript{$2$}} for time point \emph{z}. After calculating \emph{AARE\textsubscript{$2b+$\emph{5}}}, SALAD trains an LSTM model with the three most recent AARE values, i.e., \emph{A\textsubscript{$2b+$\emph{3}}}, \emph{A\textsubscript{$2b+$\emph{4}}}, \emph{A\textsubscript{$2b+$\emph{5}}} to replace \emph{M\textsubscript{$2$}}, and then uses this model to predict the next AARE value, denoted by $\widehat{\emph{A\textsubscript{$2b+$\emph{6}}}}$ (see lines 7 to 9). When time advances to \emph{$2b+6$}, the same procedure (i.e., lines 7 to 9) repeats again.

At time point \emph{$2b+7$} (i.e., \emph{$t=2b+7$}), SALAD is ready to detect anomalies since it has calculated three AARE values, which is sufficient to calculate the detection threshold \emph{thd\textsubscript{AARE}} based on Equation 5.
\begin{equation}
thd\textsubscript{\emph{AARE}}=\mu\textsubscript{\emph{AARE}}+3\cdot\sigma\textsubscript{\emph{AARE}}
\end{equation}
where $\mu\textsubscript{\emph{AARE}}=\frac{1}{t-2b-4}\cdot \sum_{w=2b+5}^{t} \emph{AARE\textsubscript{w}}$, and standard deviation $\sigma\textsubscript{AARE}$ can be derived as below. 
\begin{equation}
\sigma\textsubscript{\emph{AARE}}=\sqrt{\frac{\sum_{w=2b+5}^{t}{(AARE\textsubscript{\emph{w}}-\mu\textsubscript{\emph{AARE}})^2}}{t-2b-4}}
\end{equation} 

If \emph{AARE\textsubscript{$2b+7$}} is smaller than or equal to \emph{thd\textsubscript{AARE}} (i.e., line 12 holds), it means that \emph{A\textsubscript{$2b+7$}} is similar to previous AARE values. In this case, SALAD does not consider the corresponding data point (i.e., \emph{v\textsubscript{$2b+7$}}) anomalous, and it keeps using the current LSTM model (i.e., \emph{M\textsubscript{$2$}}) to predict the next AARE value, i.e., $\widehat{\emph{A\textsubscript{$2b+8$}}}$ (Please see lines 12 and 13 of Fig. 2).

However, if \emph{AARE\textsubscript{$2b+7$}} is larger than \emph{thd\textsubscript{AARE}} (see line 14), implying that either the data pattern of the AARE series has changed or an anomaly might happen, SALAD trains an LSTM model by taking the three most recent AARE points (i.e., \emph{A\textsubscript{$2b+$\emph{4}}}, \emph{A\textsubscript{$2b+$\emph{5}}}, \emph{A\textsubscript{$2b+$\emph{6}}}) as the training data. After that, SALAD uses the new trained LSTM model to re-predict $\widehat{\emph{A\textsubscript{$2b+7$}}}$ and then re-calculate \emph{AARE\textsubscript{$2b+7$}} (see lines 15 and 16).

If the new \emph{AARE\textsubscript{$2b+7$}} is smaller than or equal to \emph{thd\textsubscript{AARE}} (see line 17), SALAD considers that the data pattern of the AARE series has slightly changed and that \emph{v\textsubscript{$2b+7$}} is not anomalous. In this case, SALAD replaces \emph{M\textsubscript{$2$}} with this newly trained LSTM model so as to adapt to the pattern change (see line 18). This double-check approach (i.e., retraining LSTM and re-comparing the newly calculated \emph{AARE\textsubscript{$2b+7$}} with \emph{thd\textsubscript{AARE}}) enables SALAD to adapt to minor pattern change without being too sensitive.

On the contrary, if the new \emph{AARE\textsubscript{$2b+7$}} is still larger than \emph{thd\textsubscript{AARE}} (see line 19), SALAD immediately reports data point \emph{v\textsubscript{$2b+7$}} as an anomaly since the LSTM trained with the most recent AARE values is still unable to well predict \emph{A\textsubscript{$2b+7$}}. The same detection procedure will repeat over and over again to detect any possible anomaly as time further advances.

\section{Experiment Results}
To evaluate SALAD, we compared its performance against the performance of RePAD, ReRe, GrammarViz 3.0, ADT, and ADV by conducting two experiments. Recall that RePAD \cite{lee2020repad} and ReRe \cite{lee2020rere} are two state-of-the-art real-time anomaly detection approaches. The reason why they were chosen is that, similar to SALAD, they require neither offline training nor manually determined detection thresholds. GrammarViz 3.0 \cite{senin2018grammarviz} is a statistical open-source anomaly detection approach for anomalous pattern discovery in recurrent time series. ADT and ADV are two open-source statistical-based approaches introduced by Twitter. Note that we did not evaluate luminol \cite{luminol} and Matrix Profile  \cite{yeh2016matrix} since both of them rely on human experts to further set a detection threshold or top-\emph{K} discords for determining which data points are anomalies. It would be difficult or subjective to select an appropriate detection threshold or an appropriate value for \emph{K} without trial and error since users have no knowledge of the exact number of anomalies in the target time series.

In the first experiment, a real-world open-source time series dataset called New York City Taxi demand (NYC for short) was chosen from the Numenta Anomaly Benchmark (NAB) \cite{NAB}. This dataset consists of aggregating the total number of taxi passengers in New York city into 30 minute buckets from July 2014 to January 2015. This dataset has been also used by several prior works, e.g.,\cite{VictoriaZhang}\cite{NYC}\cite{farahani2019time}, to evaluate their prediction accuracy. In the second experiment, another real-world time series dataset was chosen from Taipei Mass Rapid Transit (MRT) \cite{nakamuramerlin}\cite{yehonline}. This dataset records passenger volume at the Taipei Xindian District Office metro station in February 2016. This dataset is abbreviated as MRT in this paper.
\begin{table}[t]
  \caption{Two real-world time series datasets.}
  \label{tab:Table1}
  \begin{tabular}{cp{2.9cm}p{2cm}p{1.7cm}}
    \hline
     Name &  \centering Time Period &   \centering The total number of data points & Time Interval\\
      \hline
      \hline
    NYC &   \centering From 2014-07-01, 00:00 to 2015-01-31, 23:30 &   \centering 10320 & 30 min \\
    \hline
    MRT &   \centering From 2016-02-01, 05:00 to 2016-02-29, 01:00  &   \centering 609 & 1 hour\\
     \hline
  \end{tabular}
\end{table}

\begin{table}[t]
  \caption{Parameter setting of the two experiments.}
  \label{tab:Table2}
  \begin{tabular}{cp{2.9cm}p{2.9cm}}
    \hline
    Approach & Experiment 1 (NYC) & Experiment 2 (MRT)\\
    \hline
    \hline
    SALAD & \emph{b}=288 & \emph{b}=42\\
    \hline
    RePAD & \emph{LB} = 3  & \emph{LB} = 3\\
    \hline
    ReRe & \emph{LB} = 3  & \emph{LB} = 3\\
    \hline
    GrammarViz 3.0 & \emph{PS}=4, \emph{AS}=4,\emph{WS}=49 & \emph{PS}=4, \emph{AS}=4,\emph{WS}=22\\
    \hline
    ADT &  \emph{max\_anoms}=0.02  & \emph{max\_anoms}=0.02\\
    \hline
    ADV & \emph{max\_anoms}=0.02, \emph{period}=48 & \emph{max\_anoms}=0.02, \emph{period}=21\\
    \hline
  \end{tabular}
\end{table}
TABLE I provides key information about the two datasets. The interval time between data points in the NYC dataset is 30 minutes, whereas the interval time in the MRT dataset is one hour. Note that there is no record between 2AM and 5AM every day in the MRT dataset. Hence, there are only 21 data points per day, and the total number of data points is therefore 609 (=21$*$29).

TABLE II shows all parameter settings for each of the six abovementioned approaches. When SALAD is employed, we only need to set a value for parameter \emph{b}. Recall that  \emph{b} indicates the number of data points used to train a simple LSTM model in the conversion algorithm. Since the data pattern of NYC is more complicated than that of MRT, the value of  \emph{b} in the first experiment is set to be larger than that in the second experiment. Recall that both RePAD and ReRe are based on short-term historic data points to detect anomalies, we followed the setting used in \cite{lee2020repad}\cite{lee2020rere}\cite{lee2021repad} (i.e., value 3 for the Look-back parameter, abbreviated as \emph{LB} in TABLE II) for these two approaches in both experiments.

On the other hand, three parameters are required to set before  we can use GrammarViz 3.0. These three parameters are \emph{PS} (which indicates \emph{PAA size}), \emph{AS} (i.e., \emph{alphabet size}), and \emph{WS} (i.e., \emph{window size}). We followed the default settings suggested by \cite{GrammarViz3.0} to set both \emph{PS} and \emph{AS} to 4 in both experiments. Furthermore, parameter \emph{WS} was intentionally set to 49 and 22 in the first and second experiments, respectively, since these two values enable GrammarViz 3.0 to achieve its best detection accuracy. 

For both ADT and ADV, it is required to set parameter \emph{max\_anoms} (which indicates the maximum number of anomalies to detect as a percentage of the time series). We followed the setting mentioned in \cite{Twitter} and set \emph{max\_anoms} as 0.02. Furthermore, it is also a requirement to set parameter \emph{period} for ADV. In the first experiment, we set this parameter to 48 (because totally 48 data points were collected per day). In the second experiment, we set this parameter to 21 since totally 21 data points were collected per day. Note that these values were intentionally set since they enable ADV to achieve its best detection accuracy.

TABLE III lists all LSTM hyperparameters used by SALAD, RePAD, and ReRe. Note that all LSTM models utilized by these approaches always have a simple structure: only one hidden layer with only 10 hidden units. With respect to epoch (which is defined as one forward pass and one backward pass of all the training data), it is clear that too many epochs might overfit the training data, whereas too few epochs may underfit the training data. To address this issue, whenever SALAD needs to train/retrain LSTM model, just like RePAD and ReRe, Early Stopping \cite{Early} is employed, which is an approach to automatically determine the number of epochs for preventing LSTM from overfitting the training data. In this paper, Early Stopping always chooses a number between 1 and 100 for LSTM \emph{M\textsubscript{$1$}}, and a number between 1 and 50 for LSTM \emph{M\textsubscript{$2$}}. We can see that the domain range for \emph{M\textsubscript{$1$}} is bigger than that for \emph{M\textsubscript{$2$}}  since the number of data points used to train \emph{M\textsubscript{$1$}} is more than the number of data points used to train \emph{M\textsubscript{$2$}}. On the other hand, for RePAD and ReRe, Early Stopping always chooses a number between 1 and 50 since \emph{LB} is both set to 3 in RePAD \cite{lee2020repad} and ReRe \cite{lee2020rere}.

We used a laptop running MacOS 10.15.6 with 2.6 GHz 6-Core Intel Core i7 and 16GB DDR4 SDRAM to conduct the two experiments. The purpose is to evaluate the performance efficiency of all the six approaches when they are separately deployed on a commodity machine like the abovementioned laptop. In order to evaluate the prediction accuracy for each of these approaches, we calculate precision, recall, and F-score for each of the six approaches in both experiments. Recall that precision = $\frac{TP}{TP+FP}$ and  recall=$\frac{TP}{TP+FN}$ where \emph{TP} is the total number of true positives, \emph{FP} is the total number of false positives, and \emph{FN} is the total number of false negatives. F-score is defined as the weighted harmonic mean of the precision as follows:
\begin{equation}
\emph{\emph{F-score}}=2\cdot \frac{precision\cdot recall}{precision+recall}
\end{equation}
The F-score reaches the best value, meaning perfect precision and recall, at a value of 1. The worst F-score would be a value of 0, implying the lowest precision and the lowest recall.

Note that in this paper we did not employ traditional point-wise metrics to measure precision, recall, and F-score since it is acceptable for an algorithm to trigger an alert if the delay of detecting an anomaly is not too long. It is also nice if an algorithm is able to proactively detect an anomaly before the anomaly happens.  Therefore, we adopted and revised the evaluation method proposed by \cite{ren2019time} to provide appropriate and fair comparison. 

More specifically, if any anomaly lasting for a period from time point \emph{s} to time point \emph{e} can be detected by an approach within the time period from time point \emph{s$-$X} to time point \emph{e$+$X} (where \emph{X} is a small number of time points), we say that this approach is able to detect this anomaly, and the time period from \emph{s$-$X} to \emph{e$+$X}  is called Valid Detection Period in the rest of this pape, where \emph{X} $<$ \emph{s} and \emph{e}. According to \cite{ren2019time}, value 3 is suggested for \emph{X} if the data points in a dataset are collected every hour. Since the data points in the MRT dataset were collected every hour, so we set \emph{X} to be 3 in the second experiment. On the other hand, the data points in the NYC dataset were collected every half hour, we set \emph{X} to be 6 (i.e., 2$*$3) in the first experiment.

\begin{table}[t]
  \caption{LSTM hyperparameters.}
  \label{tab:Table3}
  \begin{tabular}{p{3cm}p{3cm}cc}
    \hline
      \centering Approach &   \centering SALAD & RePAD & ReRe\\
    \hline
    \hline
      \centering The number of hidden layers &   \centering 1 & 1 & 1 \\
    \hline
      \centering The number of hidden units in a hidden layer &   \centering 10  & 10 & 10\\
     \hline
      \centering The number of epochs &   \centering 1$\sim$100 for LSTM \emph{M\textsubscript{$1$}}, 1$\sim$50 for LSTM \emph{M\textsubscript{$2$}}  & 1$\sim$50 & 1$\sim$50\\
    \hline
  \end{tabular}
\end{table}
\begin{figure*}[ht]
  \centering
  \includegraphics[width=1\linewidth]{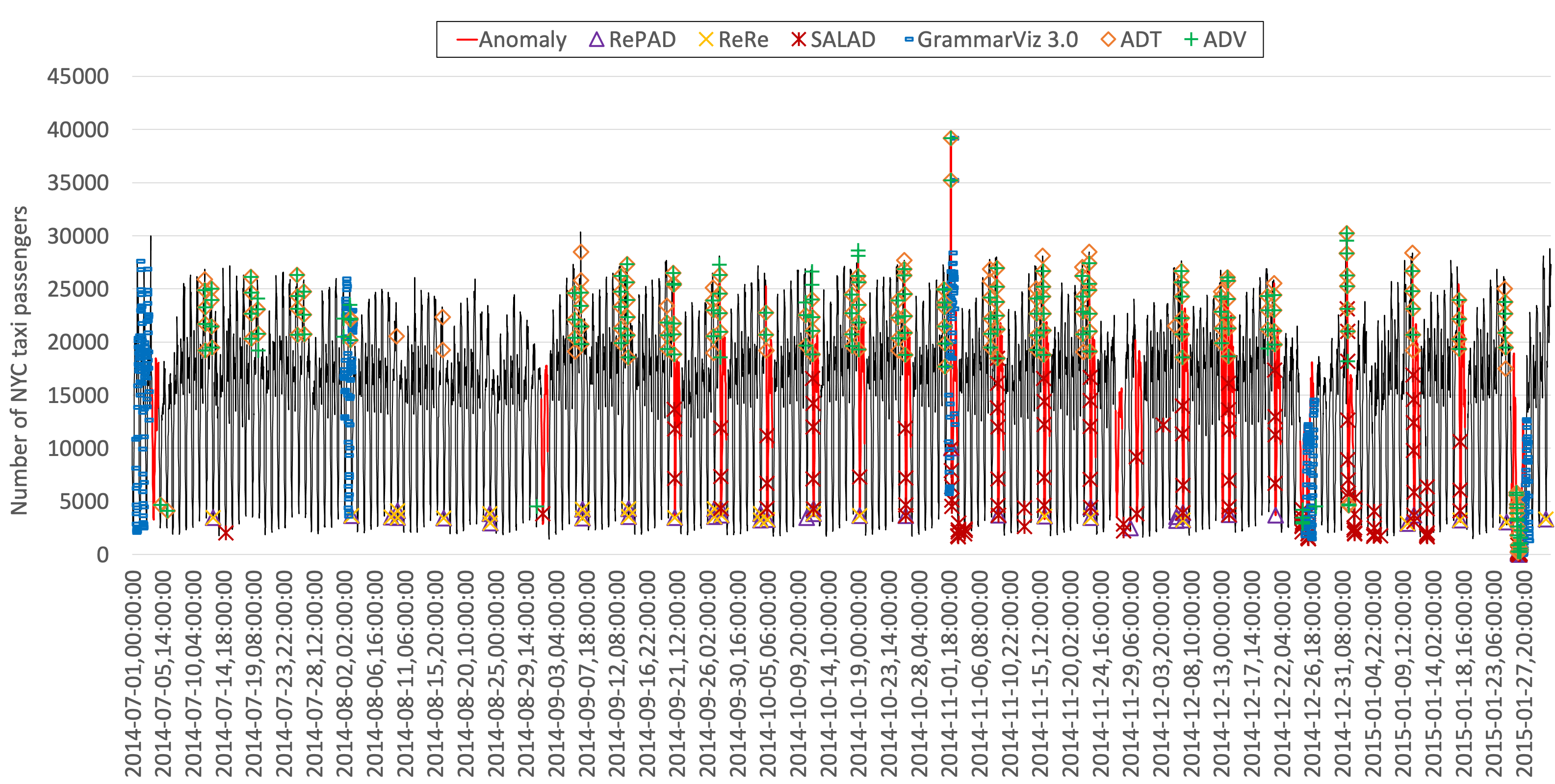}
  \caption{The detection results of all the six approaches on the NYC dataset in the first experiment.}
  \label{fig:Figure3}
\end{figure*}
\begin{figure}[ht]
  \centering
  \includegraphics[width=0.5\textwidth]{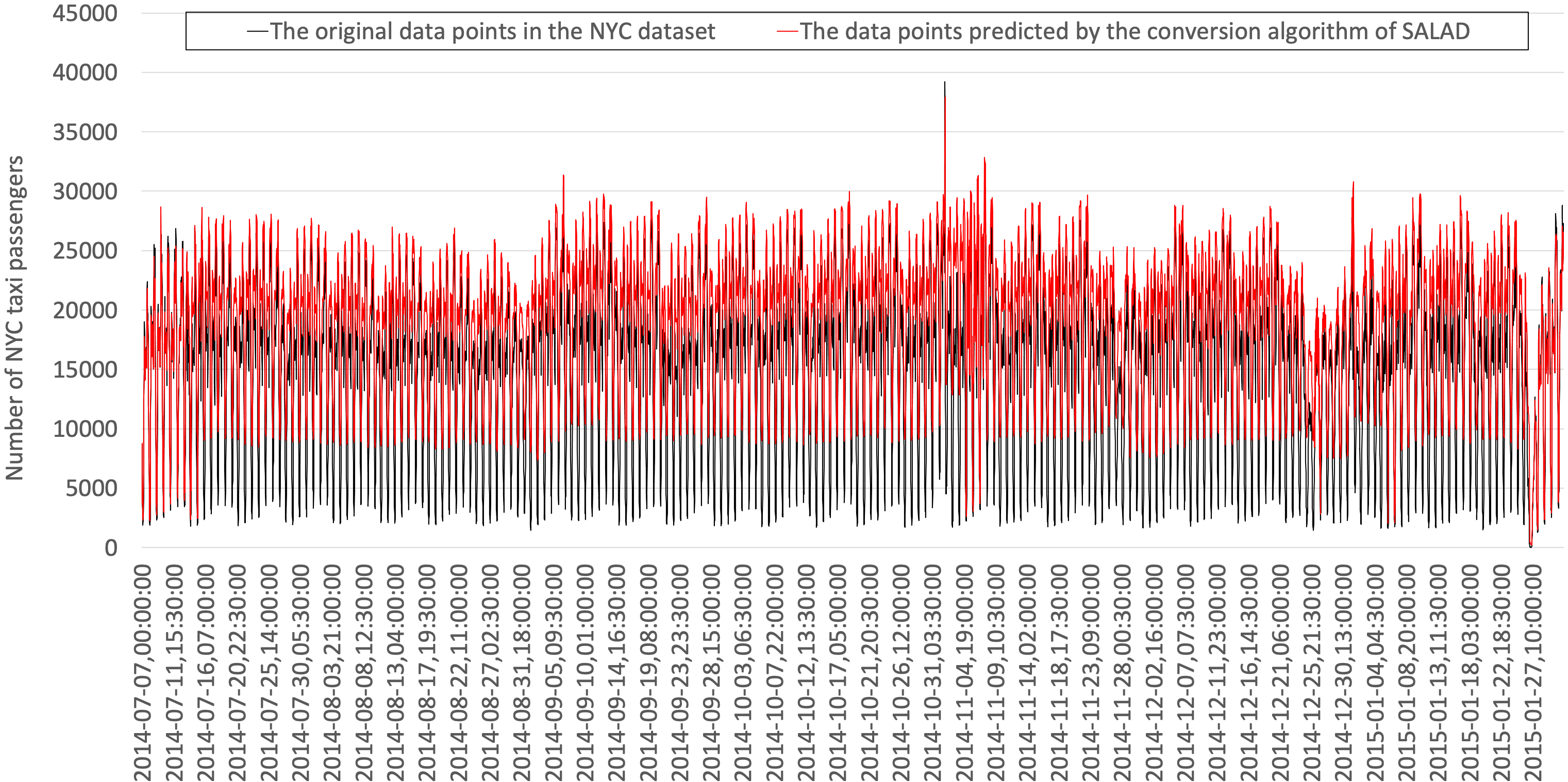}
  \caption{The original data points in the NYC dataset versus the data points predicted by the conversion algorithm of SALAD.}
  \label{fig:Figure4}
\end{figure}
\begin{figure}[ht]
  \centering
  \includegraphics[width=0.5\textwidth]{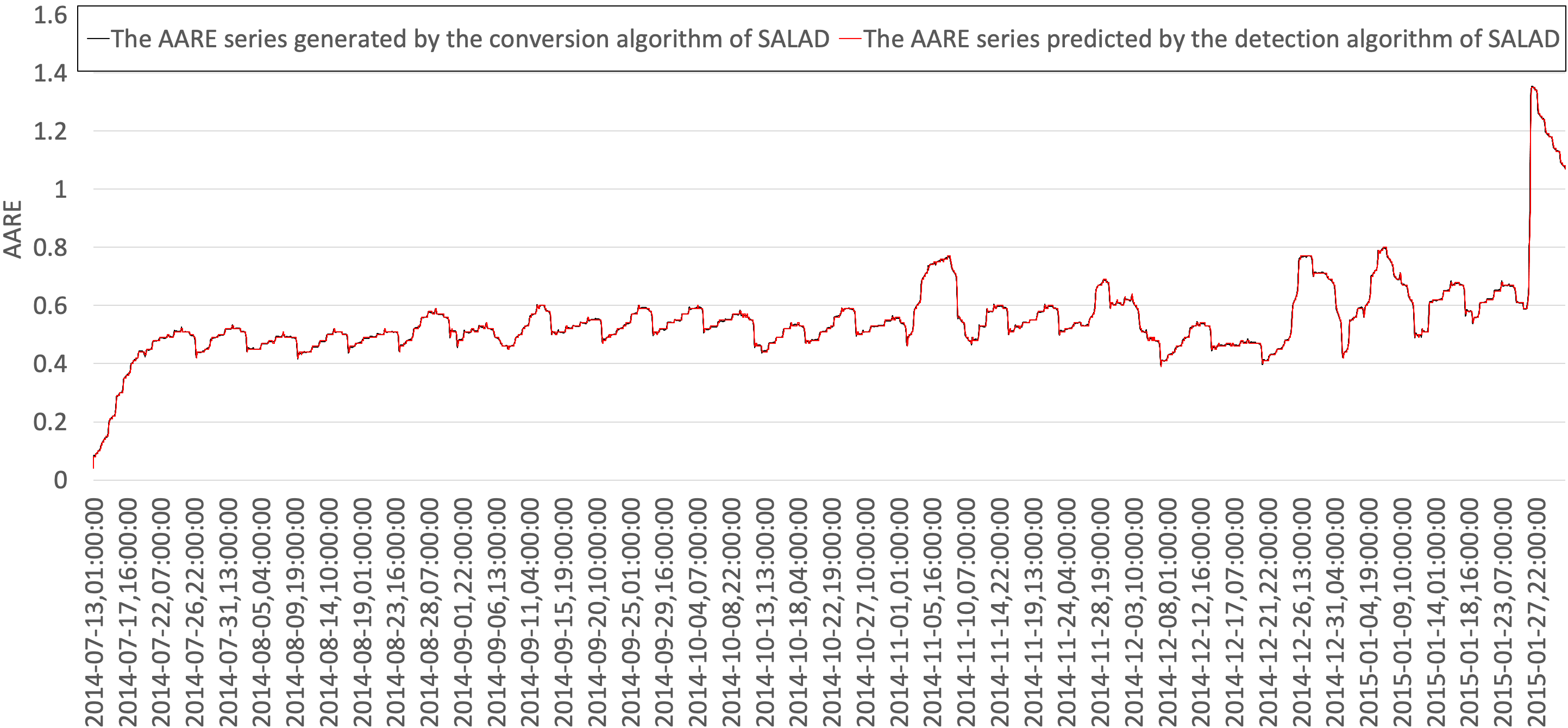}
  \caption{The AARE series generated by the conversion algorithm of SALAD versus the AARE series predicted by the detection algorithm of SALAD in the first experiment.}
  \label{fig:Figure5}
\end{figure}
\begin{figure*}[ht]
  \centering
  \includegraphics[width=0.8\textwidth]{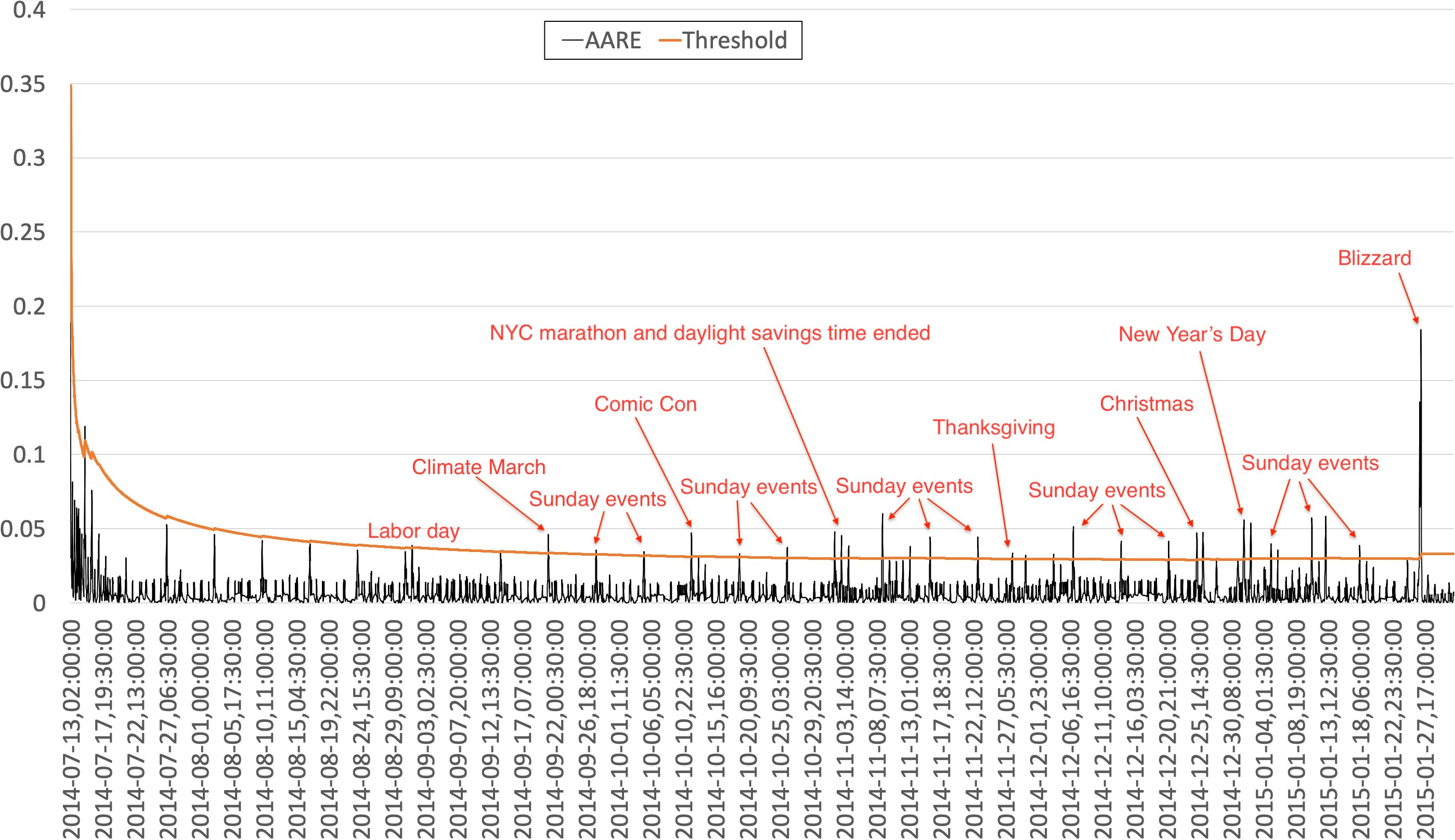}
  \caption{The AARE values generated by the detection algorithm of SALAD versus the self-adaptive detection threshold in the first experiment.}
  \label{fig:Figure6}
\end{figure*}

\subsection{Experiment 1}
As mention earlier, dataset NYC was used to evaluate all the six approaches in the first experiment. This dataset contains five anomalies identified by \cite{VictoriaZhang}\cite{NYC}:
\begin{enumerate}
\item The NYC marathon and Daylight Saving Time Ended (Nov. 2, 2014) \cite{Marathon}\cite{Clock}
\item Thanksgiving (Nov. 27, 2014) \cite{Thanksgiving}
\item Christmas (Dec. 25, 2014)
\item New Year's Day (Jan. 1, 2014)
\item Blizzard (Jan. 26-27, 2015) \cite{Blizzard}
\end{enumerate}
In addition, we also noticed other anomalies (i.e., obvious taxi demand change) that were caused by the following holidays and events/activities in NYC on Sundays:
\begin{enumerate}
\item Independence Day (July 4, 2014)
\item Labor Day (Sept. 1, 2014)
\item Climate March (Sept. 21, 2014) \cite{Climate}
\item New York Comic Con (Oct. 12, 2014) \cite{Comic}
\item Football match (Nov. 23, 2014) \cite{Football}
\item Millions March NYC (Dec. 13-14, 2014) \cite{March}
\item MLK day (Jan. 19, 2015) \cite{MLK}
\end{enumerate}

Fig. 3 illustrates each data point in the NYC dataset and the detection results of all the six approaches. All the abovementioned anomalies were highlighted in red. Apparently, each of these anomalies lasted for a period of time. TABLE IV lists the detection accuracy of all the approaches on the dataset.

As we can see from Fig. 3, both RePAD and ReRe are able to detect several anomalies, including the NYC marathon and daylight savings time ended day, the blizzard, and some Sunday events. The key reason is that both RePAD and ReRe utilize short-term historical time series data points to predict future data-point values and utilize long-term self-adaptive thresholds to detect anomalies. However, both of them repeatedly made a number of false positives since, as mentioned earlier, they are unable to learn any recurrent data pattern in a recurrent time series. Even though ReRe was designed to reduce the false positive problem of RePAD, ReRe fails to detect more anomalies than RePAD did in this dataset. This is why ReRe has a slightly lower F-score than RePAD.

Recall that we intentionally set three required parameters to GrammarViz 3.0 in order to enable this approach for achieving its best prediction accuracy on the NYC dataset. Even so, GrammarViz 3.0 can only detect three anomalies: The NYC marathon and daylight savings time ended on Nov. 2nd 2014,  Christmas on Dec. 25th 2014, and the blizzard between Jan. 26th and 27th 2015. As compared with all the other approaches, GrammarViz 3.0 has the lowest F-score. Recall that both ADT and ADV are statistical-based approaches and they require to analyze the entire NYC dataset before producing their detection results. It is clear that both approaches are able to detect a number of anomalies, but they also introduce some false positives and false negatives.

Among all the approaches, it is clear that SALAD achieves the highest precision, recall, and F-score, implying that SALAD provides the best prediction accuracy. The reason behind this is mainly due to the employed conversion and detection algorithms. Recall that the conversion algorithm of SALAD aims to convert a complex time series into a less complex AARE series by predicting a value for each future data point, measuring the difference between each pair of predicted and actual data points, and calculating the corresponding AARE values. As we can see from Fig. 4 that not all data points predicted by the conversion algorithm of SALAD perfectly match the corresponding real data points but we can see that the AARE series generated by the conversion algorithm (see the black line in Fig. 5) is much less complex than the original NYC time series (see the black line in Fig. 4). Such a conversion facilitates the detection algorithm of SALAD to learn the dataset and make correct predictions with a simple LSTM model.

Due to the above reason, we can see from Fig. 5 that the AARE series predicted by the detection algorithm of SALAD is similar to the AARE series generated by the conversion algorithm. In fact, all the resulting AARE values between the two AARE series are presented in Fig. 6. Apparently, most values are quite low, and those AARE values that are higher than the dynamically self-adaptive detection threshold are considered anomalies by SALAD. Since the threshold considers a good overview of the all previous AARE values, it is able to capture all anomalies (except the thanksgiving one on Nov. 27th, 2014) without generating many false positives or false negatives. 

Since SALAD, RePAD, and ReRe are designed to be real-time anomaly detection approaches, it is also important and necessary to compare their individual time efficiency of detecting anomalies. Note that GrammarViz 3.0, ADT, and ADV are statistical-based and offline-based approaches, such an evaluation and comparison cannot be applied to them. TABLE V shows the average time taken by SALAD, RePAD, and ReRe to determine if a data point is anomalous or not. Note that the detection time for each data point includes both the corresponding LSTM retraining time (if retraining is necessary) and the corresponding detection time. Apparently, RePAD consumes the shortest average detection time with the lowest standard deviation. This is because that RePAD is based on a simple LSTM model trained with short-term historical data points. The average detection time of ReRe is slighter higher than RePAD since ReRe employs two simple LSTM models for anomaly detection. 

The average detection time required by SALAD is around two seconds longer than those of RePAD and ReRe. The key reason is that SALAD needs to convert the NYC time series into the corresponding AARE series and then conducts the detection algorithm to detect anomalies. During these two phases, SALAD needs to train its simple LSTM model and might need to re-train its LSTM model in order to adapt to pattern changes. Nevertheless, we can see that SALAD is still time-efficient and lightweight even though it is just deployed on a regular laptop. The short response time helps trigger immediate reaction or countermeasures for handling detected anomalies.

\begin{table}[ht]
  \caption{detection accuracy on the NYC dataset.}
  \begin{center}
  \label{tab:Table4}
  \begin{tabular}{cccc}
    \hline
    Approach &Precision & Recall & F-score\\
    \hline
    \hline
    SALAD & 0.978 & 0.957 & 0.967\\
    \hline
    RePAD & 0.973  & 0.739 & 0.840\\
    \hline
    ReRe & 0.976  & 0.727 & 0.833\\
    \hline
    GrammarViz 3.0 & 0.548 & 0.222 & 0.316\\
    \hline
    ADT &  0.897  & 0.826 & 0.860\\
    \hline
    ADV & 0.911 & 0.870 & 0.890\\
    \hline
  \end{tabular}
  \end{center}
\end{table}
\begin{table}[ht]
  \caption{average detection time on the NYC dataset.}
  \begin{center}
  \label{tab:Table5}
  \begin{tabular}{ccc}
    \hline
    Approach &Average Detection Time & Standard Deviation\\
    \hline
    \hline
    SALAD & 2.072 sec & 4.145 sec\\
    \hline
    RePAD & 0.021 sec & 0.025 sec\\
    \hline
    ReRe & 0.041 sec & 0.047 sec\\
    \hline
  \end{tabular}
  \end{center}
\end{table}

\subsection{Experiment 2}

As stated earlier, we used the MRT dataset in the second experiment. This dataset is only one month long, and all its data points are illustrated in Fig. 7 (see the black line). This dataset contains only one anomaly. The anomaly is from Feb. 6th to Feb. 14th (i.e., 9 days) during which is Chinese New Year in 2016. The reason why this period was considered as an anomaly is due to the fact that the passenger volume taking Taipei MRT in this time period was apparently less than the rest days in the same month.

\begin{figure*}[ht]
  \centering
  \includegraphics[width=0.8\linewidth]{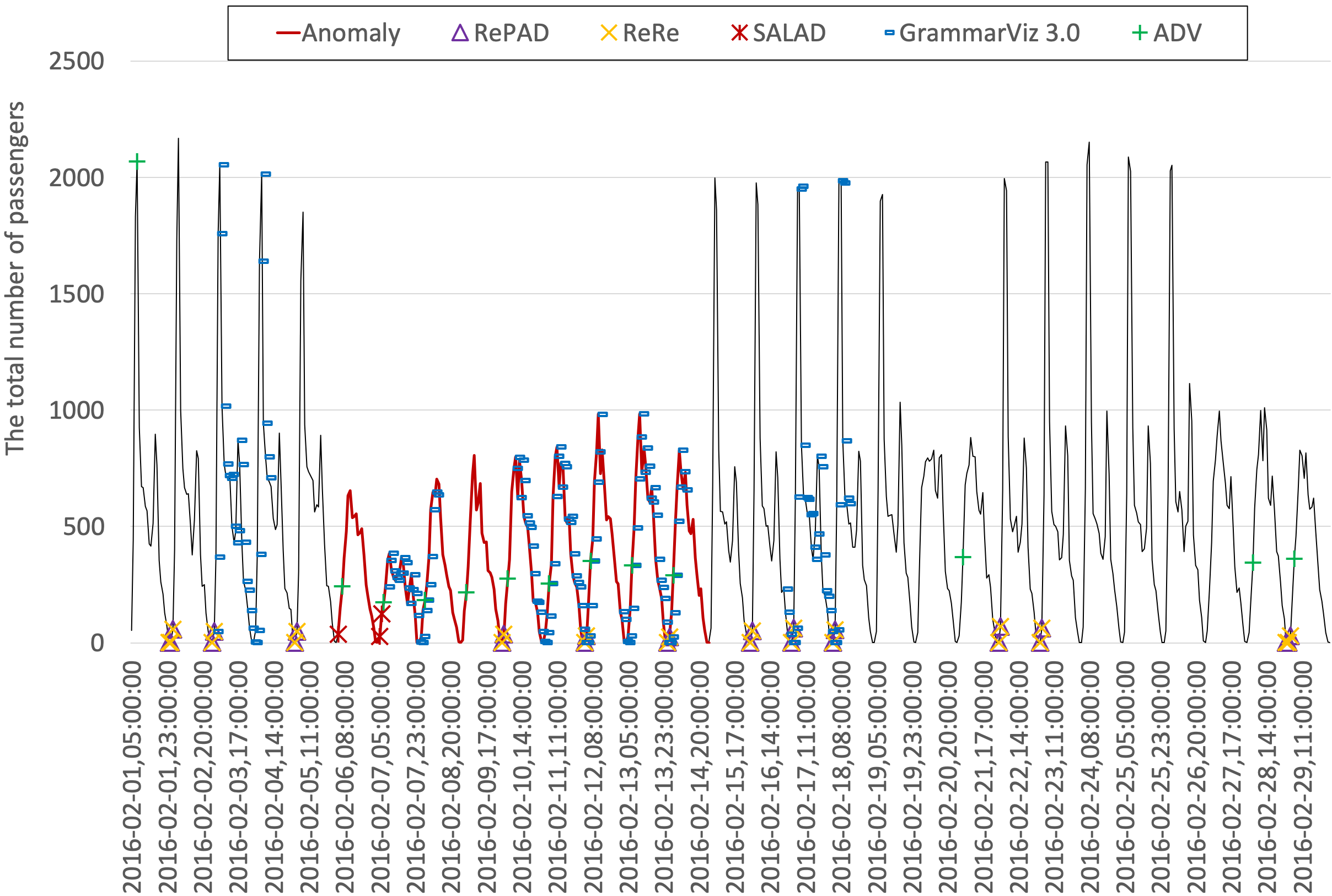}
  \caption{The detection results of all the approaches (except ADT) on the MRT dataset in the second experiment.}
  \label{fig:Figure7}
\end{figure*}
\begin{figure}[ht]
  \centering
  \includegraphics[width=0.9\linewidth]{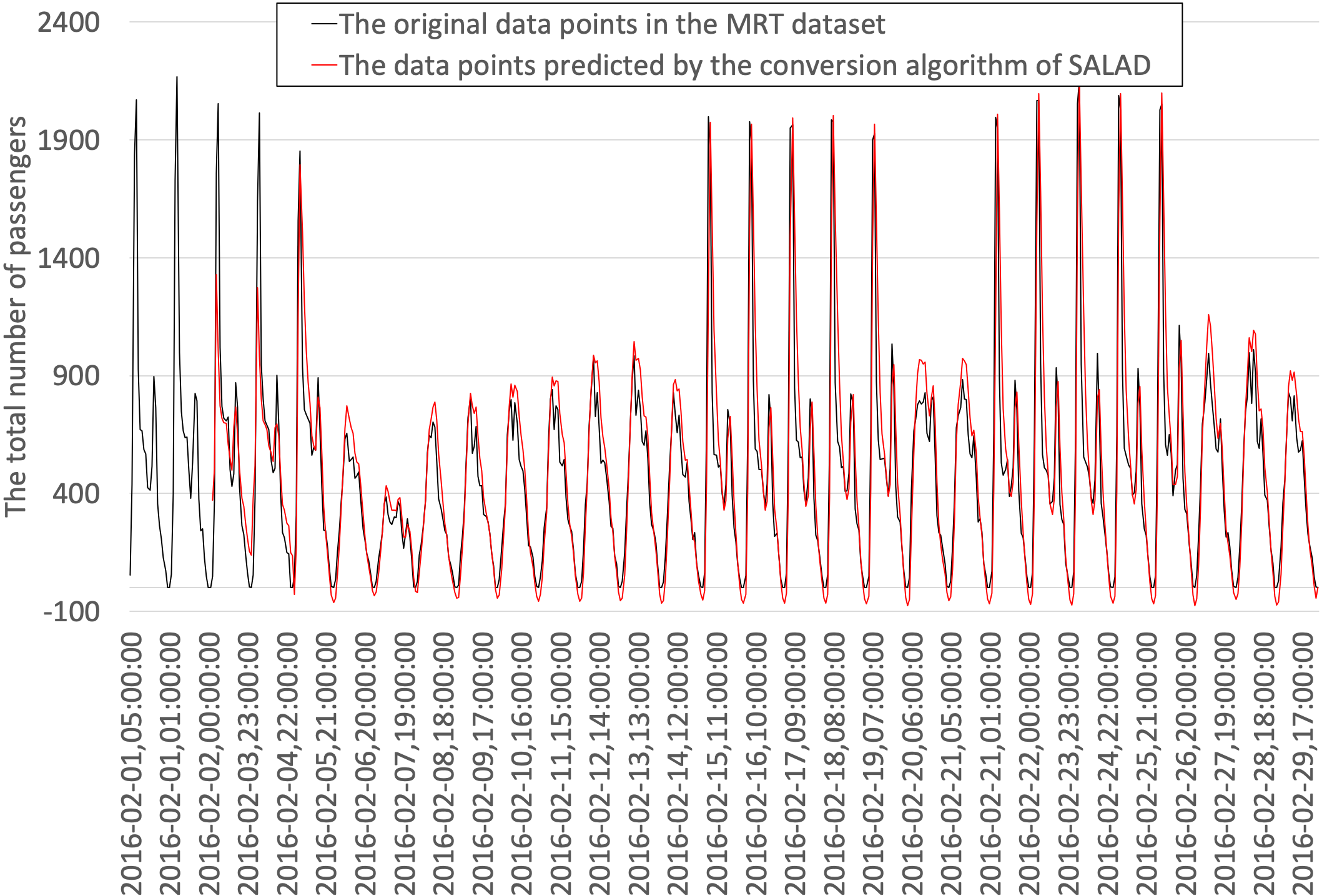}
  \caption{The original data points in the MRT dataset versus the data points predicted by the conversion algorithm of SALAD.}
  \label{fig:Figure8}
\end{figure}
\begin{figure}[ht]
  \centering
  \includegraphics[width=0.9\linewidth]{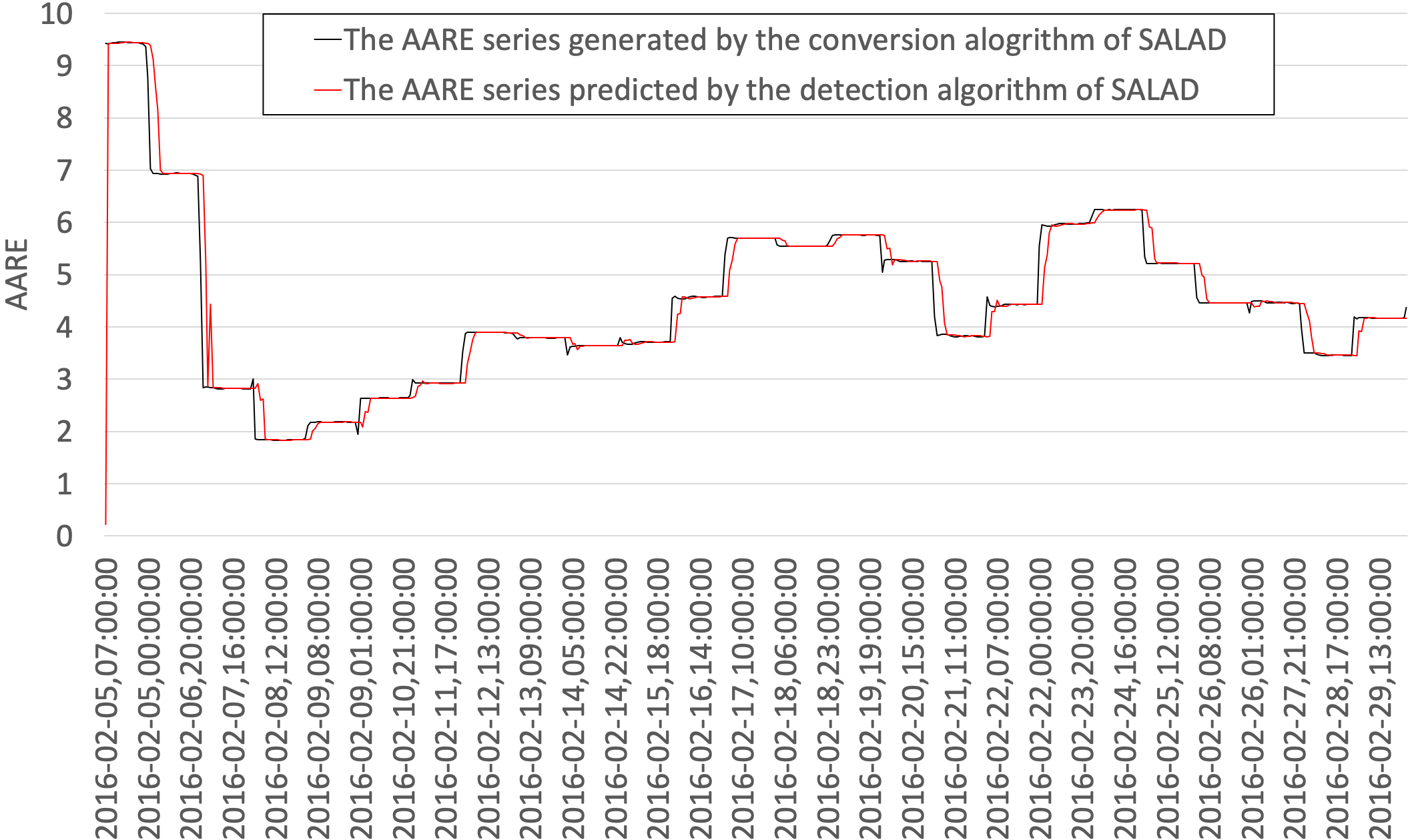}
  \caption{The AARE series generated by the conversion algorithm of SALAD versus the AARE series predicted by the detection algorithm of SALAD in the second experiment.}
  \label{fig:Figure9}
\end{figure}
\begin{figure}[ht]
  \centering
  \includegraphics[width=0.9\linewidth]{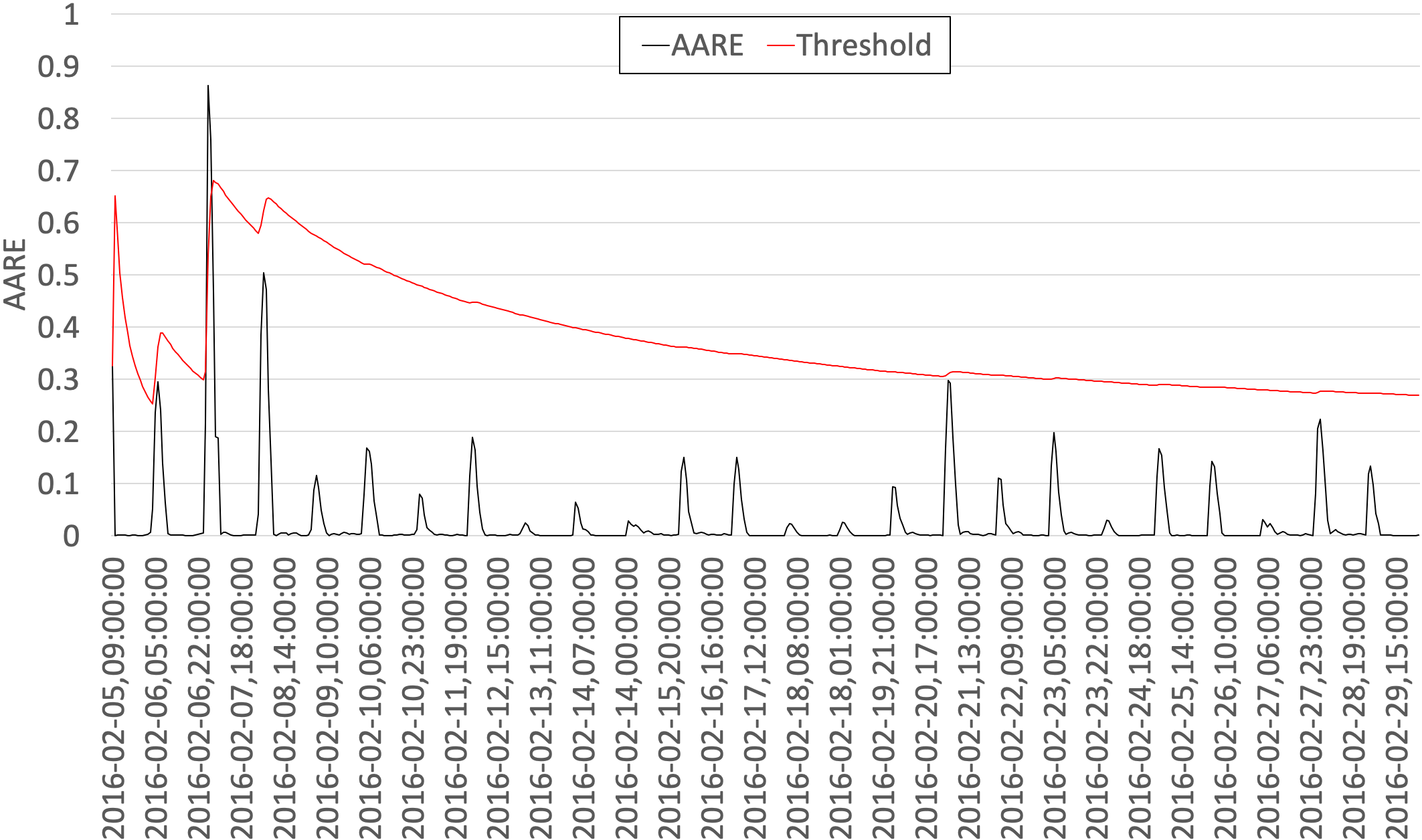}
  \caption{The AARE values generated by the detection algorithm of SALAD versus the self-adaptive detection threshold in the second experiment.}
  \label{fig:Figure10}
\end{figure}

Fig. 7 plots the detection results of all the approaches on this dataset except ADT. In the second experiment, ADT was unable to execute since it requires a dataset to be about four times longer than the length of MRT. TABLE VI also lists the detection accuracy of all the five approaches on the dataset. According to the defined Valid Detection Period, all of the approaches are able to detect the Chinese-New-Year anomaly. However, as we can see from Fig. 7, it is clear that SALAD is the first one that detects the anomaly, followed by ADV, GrammarViz 3.0, RePAD, and ReRe. 

In addition, SALAD outperforms the other four approaches when it comes to precision, recall, and of course F-score. SALAD did not make any false positives or false negatives. Therefore, its F-score reaches the best value of 1. The reason for such a great detection performance is as follows: SALAD successfully converts the MRT time series (as shown in Fig. 8) into a less complex AARE series (as shown in Fig. 9). Such an AARE series enables SALAD to easily predict using its simple LSTM structure without having to use a more complicated LSTM structure. As we can see from Fig. 10, the AARE values calculated by the detection algorithm of SALAD together with the self-adaptive detection threshold enable SALAD to easily detect the Chinese-New-Year anomaly. It is also clear that the AARE values associated with all other data points at different time points did not exceed the detection threshold, so those data points were not considered as anomalous by SALAD.  

From Fig. 7 and TABLE VI, we can see that RePAD and ReRe have the same results in F-score, and that they generated the same number of false positives. Apparently, RePAD and ReRe are unable to perfectly learn the recurrent data pattern of the MRT dataset by using short-term historical data points (i.e., \emph{LB} = 3 \cite{lee2021repad}) to determine the anomalousness of each data point. 

Recall that the Chinese-New-Year anomaly lasted for 9 days in the entire MRT dataset, meaning that around 31\% of data points are anomalous. This kind of the dataset introduces some detection difficulty for GrammarViz 3.0 since this approach tends to consider infrequent and irregular data points as anomalies. Due to this reason, GrammarViz 3.0 made some false positives even though it was able to detect the Chinese-New-Year anomaly in the valid detection period. 

When ADV was employed, we can see from Fig. 7 that it also considered some weekends (e.g., 21th, 28th, and 29th of the February) as anomalies since this approach is statistical based. It seems unable to differentiate the data pattern in these weekends and that in the Chinese-New-Year period. 

TABLE VII shows the average detection time of SALAD, RePAD, and ReRe on the MRT dataset. Due to the same reason mentioned earlier, the average detection time of SALAD is slightly higher than those of RePAD and ReRe. Nevertheless, SALAD still offers time-efficient and lightweight detection performance due to its ability of converting the MRT dataset into a less complex AARE series and its ability to predict AARE values by using a simple LSTM model, rather than employing deep network structures.
\begin{table}[ht]
  \caption{detection accuracy on the MRT dataset.}
  \begin{center}
  \label{tab:Table6}
  \begin{tabular}{cccc}
    \hline
    Approach &Precision & Recall & F-score\\
    \hline
    \hline
    SALAD & 1 & 1 & 1\\
    \hline
    RePAD & 0.913  & 1 & 0.955\\
    \hline
    ReRe & 0.913  & 1 & 0.955\\
    \hline
    GrammarViz 3.0 & 0.762 & 1 & 0.865\\
    \hline
    ADT &  N/A  &N/A & N/A\\
    \hline
    ADV & 0.979 & 1 & 0.989\\
    \hline
  \end{tabular}
  \end{center}
\end{table}
\begin{table}[ht!]
  \caption{average detection time on the MRT dataset.}
  \begin{center}
  \label{tab:Table7}
  \begin{tabular}{ccc}
    \hline
    Approach &Average Detection Time & Standard Deviation\\
    \hline
    \hline
    SALAD & 0.109 sec& 0.174 sec\\
    \hline
    RePAD & 0.031 sec& 0.064 sec\\
    \hline
    ReRe & 0.065  sec& 0.100 sec\\
    \hline
  \end{tabular}
  \end{center}
\end{table}
\section{Conclusion and Future Work}
In this paper, we have introduced SALAD, which is designed to detect anomalies in a recurrent time series on the fly in real time. SALAD does not need to go through an offline training process to pre-train a learning model, nor require human experts to tune several parameters or determine a detection threshold in advance. By converting a complex time series into a less complex AARE series, SALAD is able to effectively and efficiently conduct anomaly detection using a simple network structure without having to employ a deep network structure, thereby reducing computation complexity and enabling itself to be deployed and executed on a commodity machine. Based on the self-adaptive detection threshold that is automatically derived and updated at each time point, SALAD does not suffer the issue of top-\emph{K} discords or requires users to decide an appropriate detection threshold beforehand.  

The experiments based on two real-world open-source time series datasets demonstrate that SALAD provides better detection accuracy than the other five state-of-the-art anomaly detection approaches. In addition, SALAD does not introduce significant computation overhead, which makes it a practical solution to be deployed on a commodity machine in real scenarios.

As future work, we would like to study an approach that is able to automatically determine an appropriate value for parameter \emph{b} such that it suits any given recurrent time series. 
In addition, we plan to introduce an approach that is able to automatically recognize whether a given time series is recurrent or not on the fly and then use an appropriate anomaly detection algorithm. Furthermore, we would also like to extend and deploy SALAD on the eX\textsuperscript{3} HPC cluster \cite{eX3} to further handle large-scale time series in a parallel and distributed way. by referring to \cite{lee2015hybrid}\cite{lin2016performance}.

\section*{Acknowledgment}
This work was supported by the project eX\textsuperscript{3} - \emph{Experimental Infra-structure for Exploration of Exascale Computing} funded by the Research Council of Norway under contract 270053 and the scholarship under project number 80430060 supported by Norwegian University of Science and Technology.

\vspace{12pt}

\end{document}